\documentclass[letterpaper, 10 pt, journal, twoside]{IEEEtran}
\usepackage{graphicx}
\usepackage{amsfonts}
\usepackage{amsmath}
\usepackage{bm}
\usepackage{booktabs}
\usepackage{multirow}
\usepackage{url}
\usepackage{xcolor}
\usepackage{caption}  
\usepackage{fontawesome}  
\ifCLASSINFOpdf
\else
\fi
\begin{document}

%
\title{\LARGE \bf
\textit{Mimir}: {Hierarchical} Goal-Driven Diffusion with Uncertainty Propagation for End-to-End Autonomous Driving
}
%
%
%

\author{Zebin Xing\textsuperscript{1}, Yupeng Zheng\textsuperscript{1}\dag, Qichao Zhang\textsuperscript{1}\ddag, \\
Zhixing Ding\textsuperscript{1,2}, Pengxuan Yang\textsuperscript{1}, Songen Gu\textsuperscript{3}, Zhongpu Xia\textsuperscript{1}, and Dongbin Zhao\textsuperscript{1} \textit{Fellow, IEEE}\\
\textsuperscript{1} Institue of Automation, Chinese Academy of Sciences \\
\textsuperscript{2} China University of Geosciences
\textsuperscript{3} Fudan University
\thanks{
        The email of Zebin Xing is {\tt\footnotesize xzebin@bupt.edu.cn}%
        {\tt\footnotesize }%
}
\thanks{Yupeng Zheng\dag~is the projector leader. Qichao Zhang\ddag~is the corresponding author.
}

}
%
%

\markboth{IEEE Robotics and Automation Letters. Preprint Version. Accepted November 2025}
{Xing \MakeLowercase{\textit{et al.}}: Goal-Driven Diffusion with Uncertainty}

%



\maketitle


\begin{abstract}
End-to-end autonomous driving has emerged as a pivotal direction in the field of autonomous systems. Recent works have demonstrated impressive performance by incorporating high-level guidance signals to steer low-level trajectory planners. However, their potential is often constrained by inaccurate high-level guidance and the computational overhead of complex guidance modules. To address these limitations, we propose Mimir, a novel hierarchical dual-system framework capable of generating robust trajectories relying on goal points with uncertainty estimation: 
(1) Unlike previous approaches that deterministically model, we estimate goal point uncertainty with a Laplace distribution to enhance robustness;
(2) To overcome the slow inference speed of the guidance system, we introduce a multi-rate guidance mechanism that predicts extended goal points in advance.
Validated on challenging Navhard and Navtest benchmarks, Mimir surpasses previous state-of-the-art methods with a 20\% improvement in the driving score EPDMS, while achieving 1.6$\times$ improvement in high-level module inference speed without compromising accuracy. The code and models will be released soon to promote reproducibility and further development. \textit{The code is available at \url{https://github.com/ZebinX/Mimir-Uncertainty-Driving}} 
\end{abstract}


%
\IEEEpeerreviewmaketitle

\section{Introduction}
%
%
%
%
\IEEEPARstart{E}{nd-to-end} autonomous driving, which directly maps sensor observations (e.g., RGB images and ego status) to driving actions, has emerged as a pivotal research frontier due to its potential for scalable deployment.

\begin{figure}[t]
  \centering
  \includegraphics[width=0.50\textwidth]{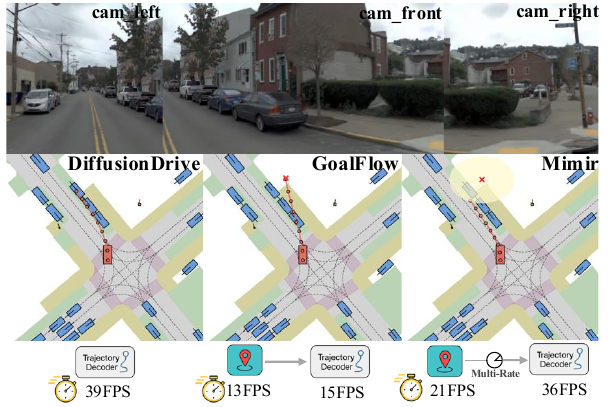}
  \caption{Compared to existing methods, Mimir assesses the uncertainty of the predicted goal points from the high-level model, enabling safer and more reliable trajectory planning. }
  \label{fig:teaser}
\vspace{-1em}
\end{figure}

Pioneering methods~\cite{Chitta2023transfuser, hu2023uniad, jiang2023vad, gao2024dream, zheng2025world4drive} regress a single-mode by constructing various scene representations and multiple auxiliary supervision tasks with raw sensors input, yet overlook the inherent variance and uncertainty of driving actions.
Recently, some approaches ~\cite{liao2025diffusiondrive, xing2025goalflow} model multi-modal action distributions with diffusion. 
DiffusionDrive~\cite{liao2025diffusiondrive} extracts anchors from trajectory vocabulary and accelerates denoising through anchor-guided diffusion, while SOTA method, GoalFlow~\cite{xing2025goalflow}, employs goal points as conditional constraints trajectory generation. 
Although GoalFlow ingeniously bridges perception and multi-modal planning via goal-dirven design, its real-world applicability faces two critical limitations:
(1) Error propagation: goal point prediction errors propagate downstream to the planning module as shown in Fig.~\ref{fig:teaser}. The goal point specifies an exact short-term future position, strongly guiding the planner. In deterministic modeling, large prediction errors may mislead the decoder into generating unsafe trajectories. 
(2) Real-time bottleneck: cascaded goal construction and trajectory generation impede real-time performance. Hierarchical frameworks~\cite{xing2025goalflow,jiang2024senna} often adopt heavy high-level models to extract effective guidance, which challenges deployment on real vehicles due to high computational cost.

To address these challenges, we propose Mimir, a novel hierarchical dual-system framework comprising:
(1) A slow guidance system that generates goal points with Laplacian-modeled uncertainty from RGB observations, ego status, and a trajectory vocabulary. It further extrapolates extended goal points and injects them into the fast planner.
(2) A fast planning system first aligns perception features and the goal point guidance in two different coordinate systems via a Guidance Injection module. Then, a diffusion-based planner generates trajectories conditioned on the sampled perception features, goal points, and associated uncertainty.

Extensive evaluations on large-scale challenging Navhard and Navtest benchmarks demonstrate that Mimir achieves SOTA performance with 20\% driving score improvement. 
Besides, leveraging the multi-rate architecture, Mimir increases 1.6$\times$ inference speed in the high-level module.

In summary, our contributions are threefold:
\begin{itemize}
    \item We propose a hierarchical dual-system framework for autonomous driving, Mimir, which achieves real-time performance by predicting extended goal points through an multi-rate guidance mechanism.
    \item We explicitly model the uncertainty of the goal point using a Laplace distribution, and incorporate this uncertainty into the trajectory planner to enable more reliable and robust trajectory generation.
    \item Our method achieves SOTA  performance on the large-scale, challenging Navhard and Navtest benchmarks, with up to a 20\% improvement in the EPDMS driving score and a 1.6$\times$ speedup in high-level module inference.
\end{itemize}

\begin{figure*}[ht]
  \centering
  \includegraphics[width=1.0\textwidth]{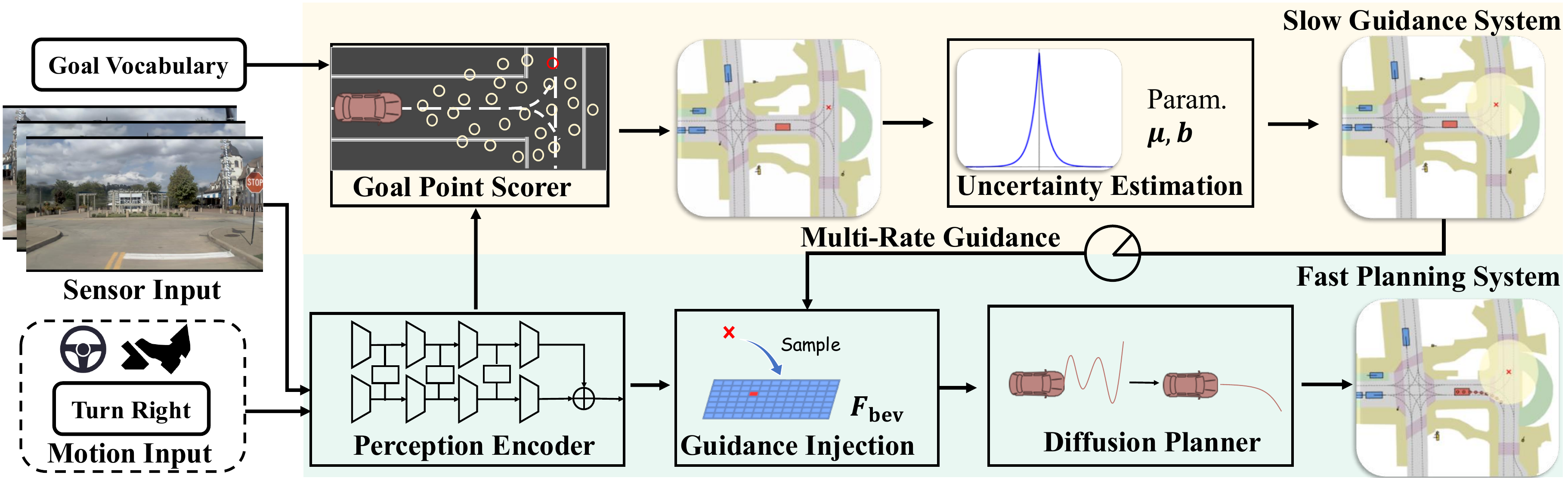}

  \caption{\textbf{Overview of the Mimir architecture.} The high-level model (orange) selects the optimal goal point from a predefined vocabulary $\mathbb{V}$ and estimates its uncertainty using a Laplace distribution to generate guidance $G=\{\bm{\mu},\bm{b}\}$. The multi-rate Guidance further predicts the extended goal point, allowing the system to operate at different rates. The low-level planner (green) fuses the guidance information with perception features through a Guidance Injection module, then generates trajectories via the diffusion planner. Mimir decouples guidance generation from trajectory planning for optimized efficiency.}
  \label{fig:main_fig}
    \vspace{-1em}
\end{figure*}

%

\section{RELATED WORK}
\subsection{End-to-End Autonomous Driving}
Earlier autonomous driving approaches~\cite{zhang2022trajgen, li2023planning, zheng2024planagent} followed a modular pipeline. End-to-end autonomous driving directly maps raw sensory inputs like camera images and LiDAR scans to driving actions via neural networks. Some approaches rely on direct trajectory regression for prediction. Methods like Transfuser~\cite{Chitta2023transfuser} advance this by effectively fusing multimodal sensor data and utilizing a GRU to output the predicted trajectory. UniAD~\cite{hu2023uniad} focuses on designing specialized transformer modules to extract critical information such as maps and visual features, ultimately also regressing the trajectory using a transformer. Pushing efficiency further, VAD~\cite{jiang2023vad} improves significantly by replacing dense BEV features with vector representations.  CIL++~\cite{xiao2023multiview} investigates multi-view fusion for trajectory prediction. Subsequently, SparseDrive~\cite{sun2024sparsedrive} builds upon this by employing sparse representations for both surrounding agents and map elements and leveraging pre-defined goal points to elicit multi-modal trajectory predictions. 

Leveraging the strong modeling capabilities of diffusion~\cite{ho2020denoising} frameworks, more recent end-to-end methods increasingly favor using a diffusion planner to model the multimodal distribution of possible trajectories. HE-Drive~\cite{wang2024he}, for instance, models trajectory distributions with a diffusion planner, aligning the predictions more closely with human driving preferences. DiffusionDrive~\cite{liao2025diffusiondrive} employs a truncated diffusion process, starting from pre-sampled anchor trajectories to generate the future path. Similarly, GoalFlow~\cite{xing2025goalflow} initially predicts future goal points and uses this guidance to steer the goal-conditioned trajectory generation performed by its flow matching~\cite{lipman2022flow} model. Although these methods attempt to model the multimodality of trajectories using the diffusion framework, they still overlook the aleatoric uncertainty of the condition. This limitation restricts the performance when the condition is inaccurate.

\subsection{Uncertainty Estimation}
Standard neural network models often lack inherent uncertainty estimation capabilities. To tackle this, certain approaches like~\cite{ramalho2020density} explicitly model uncertainty using dedicated network components, for instance, by caching the k nearest neighbors of a predicted label and employing a separate classifier to predict uncertainty. Another method~\cite{zhang2024ctrl} leverages the discrepancy between outputs from multiple diffusion processes to quantify the uncertainty associated with a predicted reward function. Alternative strategies~\cite{valdenegro2019deep} utilize model ensembles, aggregating results from multiple models to form the final prediction and estimate uncertainty. Further techniques apply Bayesian methods~\cite{welling2011bayesian} or Test-Time Augmentation~\cite{lyzhov2020greedy} to evaluate model confidence.

Several methods specifically attempt to model uncertainty for perception or planning tasks in autonomous systems. \cite{michelmore2020uncertainty} has explored Bayesian uncertainty estimation for quantitatively assessing uncertainty and evaluating its predictive quality. \cite{gu2024producing,yang2025uncad} estimates map uncertainty by modeling the distribution of points in vectorized maps. Furthermore, \cite{filos2020can} proposes a method using model ensembles to assess epistemic uncertainty in driving plans, enabling the detection of distribution shifts and facilitating recovery. However, this ensemble-based approach requires training multiple models, significantly increasing the cost of model deployment.


\section{Problem Definition}
End-to-end autonomous driving typically requires modeling perceived environmental information and observable ego-motion states without reliance on high-definition maps to generate future trajectories. 
Given history sensory inputs $\{I^\mathrm{sensor}_{-n},...,I^\mathrm{sensor}_0\}$ from time $-n$ to time $0$ (current time), where $I^\mathrm{sensor}=\{I^\mathrm{cam}, I^\mathrm{lidar}\}$, and ego-motion states $\{I^\mathrm{motion}_{-n},...,I^\mathrm{motion}_0\}$ where $I^\mathrm{motion}=\{I^\mathrm{command}, I^\mathrm{vel}, I^\mathrm{acc}\}$, the neural network needs to output a future trajectory sequence $\{O^\mathrm{traj}_{1}, ...,O^\mathrm{traj}_{T}\}$, where each trajectory point satisfies $O^\mathrm{traj}=\{x, y, heading\}$. Here, $I^\mathrm{command}$ denotes the high-level driving intention (turn left, turn right, go straight, or unknown) encoded as a one-hot vector.

This task can be formally defined as finding the maximum likelihood for the ego vehicle's future trajectory:

\begin{equation}
    \arg\max_\theta P_\theta(O^\mathrm{traj}_{1:T}\mid{I}_{-n:0}^\mathrm{sensor},{I}_{-n:0}^\mathrm{motion})
    \label{eq:task_def_straight}
\end{equation}

Several approaches~\cite{dendorfer2020goal,xing2025goalflow,jiang2024senna} employ a hierarchical framework for trajectory prediction. They first predict the high-level guidance $G$ like meta-action or goal point, then utilizes this to direct the ego vehicle's specific motion. These methodologies can be formally defined as:
\begin{equation}
    \arg\max_\theta P^\mathrm{hight}_\theta(G\mid{I}_{-n:0}^\mathrm{sensor},{I}_{-n:0}^\mathrm{motion})
    \label{eq:task_def_hier1}
\end{equation}
\begin{equation}
    \arg\max_\theta P^\mathrm{low}_\theta(O^\mathrm{traj}_{1:T}\mid{I}_{-n:0}^\mathrm{sensor},{I}_{-n:0}^\mathrm{motion},G)
    \label{eq:task_def_hier2}
\end{equation}
Where $P^\mathrm{high}_\theta(\cdot)$ represents the high-level model that captures guidance distributions. In contrast, $P^\mathrm{low}_\theta(\cdot)$ denotes the low-level model that generates trajectory distribution.

\section{METHODOLOGY}
As illustrated in Fig. \ref{fig:main_fig}, we propose Mimir, a robust and efficient fast-slow system that enhances trajectory generation quality through high-level uncertainty modeling. We begin by introducing the Perception Encoder and Goal Point Scorer to select the raw goal point initially. Next, we introduce Uncertainty Estimation of the goal point using the Laplace distribution. Following this, we present an extended goal point prediction strategy to construct our Multi-Rate Guidance. Finally, we describe how Guidance Injection is applied to effectively incorporate guidance into the Diffusion Planner.

\subsection{PERCEPTION ENCODER AND GOAL POINT SCORER}
Given the sensor inputs and ego-motion, Mimir first extracts the perception features using the Perception Encoder and then selects a raw goal point from the goal point vocabulary $\mathbb{V}$ using a Goal Point Scorer. 

\noindent \textbf{Perception Encoder.} Typical sensor inputs include image $I\in\mathbb{R}^{3\times H_{\mathrm{img}} \times W_{\mathrm{img}}}$ and LiDAR $L \in \mathbb{R}^{K\times3}$. Following Transfuser~\cite{Chitta2023transfuser}, we extract features from camera and LiDAR inputs using two separate resnet50 backbones. To enable a deep fusion between image $I$ and LiDAR $L$, a transformer is utilized at each block of the resnet, progressively integrating the backbone features into BEV feature $F_{\mathrm{bev}}$. During training, the BEV features $F_{\mathrm{bev}}$ are supervised via two heads: a map head for HD map annotations, and an agent head for bounding boxes of surrounding agents. Notably, due to the synthetic image in navsimv2~\cite{navsimv2} lacking corresponding LiDAR, a learnable feature substitutes the LiDAR input in the deployment of navsimv2.

\noindent \textbf{Goal Point Scorer.} 
We follow GoalFlow~\cite{xing2025goalflow} to model the goal point as high-level guidance, representing the ego vehicle's location at the 4-second horizon. 
Initially, we cluster the endpoints of trajectories to construct a vocabulary $\mathbb{V}=\{g_i\}^{8192}_{i=1}$ containing 8,192 candidate goal points. 
For each goal point $g_i \in \mathbb{V}$, we predict two scores: $\delta^{\mathrm{dac}}$, which measures Drivable Area Compliance (DAC), and $\delta^{\mathrm{dis}}$, which indicates the proximity between $g_i$ and the endpoint $g_\mathrm{end}$ of the ground-truth trajectory $\tau_\mathrm{gt}$ using softmax. Finally, the candidate goal point with the highest combined score is then selected as the raw goal point $g_{\mathrm{raw}}$.

\subsection{UNCERTAINTY ESTIMATION}
\label{sec:uncertainty estimation}
In previous work, hierarchical frameworks~\cite{xing2025goalflow,jiang2024senna} often employed a powerful model to plan deterministic high-level commands like meta-actions, goal points, and language instructions to control the low-level planners. However, in autonomous driving, driving behaviors are often not deterministic. Goal points are often used as guiding information in motion planning tasks~\cite{wang2022stepwise,yao2021bitrap}. Approaches like GoalFlow~\cite{xing2025goalflow} and GoalGan~\cite{dendorfer2020goal} sample the goal points from a fixed candidate set, which limits the model's ability to generalize across out-of-distribution scenarios. To address these issues, we propose using a lightweight model to capture the uncertainty in goal point guidance and extend the fixed set sampling into continuous space modeling.

The raw goal point $g_{\mathrm{raw}}$, defined by x and y coordinates, is a deterministic quantity. To model its inherent uncertainty, we transform it into a distribution over the 2D coordinate space. Common approaches~\cite{yeo2021robustness,kendall2017uncertainties} represent this uncertainty using parametric distributions such as Laplace or Gaussian centered at the predicted coordinate. Specifically, we use Laplace modeling to represent goal point distributions based on the given raw goal points. For each scenario, the x, y coordinate of the goal point are designed to satisfy the Laplace distribution, whose probability density satisfies:
\begin{equation}
    f(v_i|\mu_i,b_i)=\frac{1}{2b_i}\mathrm{exp}(-\frac{|v_i-\mu_i|}{b_i})
    \label{eq:lapalace_distribution}
\end{equation}
where \( v_i \) denotes the coordinate value (i.e., \( v_1 \) for the \(x\)-axis and \( v_2 \) for the \(y\)-axis), \( \mu_i \) is the location parameter, and \( b_i \) is the scale parameter controlling the spread of the distribution. 

As for the module structure, we first embed $g_\mathrm{raw}$ into a goal point query, which interacts with the BEV feature map $F_{\mathrm{bev}}$ through spatial cross-attention. This step allows spatial semantic context to be injected into the goal representation. Subsequently, agent-level cross-attention is performed between the goal query and $F_\mathrm{agent}$, which derives from the agent head in Perception Encoder, enabling the modeling of dynamic interactions between the goal and surrounding agents. The output is processed by a feed-forward network for non-linear transformation. Two MLP heads then predict the refined goal point \( \bm{\mu} = (\mathbf\mu_1, \mu_2) \in \mathbb{R}^2 \) and its associated uncertainty \( \bm{b} = (b_1, b_2) \in \mathbb{R}_+^2 \), modeled as the scale parameter of a Laplace distribution.

We adopt the Negative Log-Likelihood of the Laplace distribution as our loss function to independently train the $\bm{\mu}$ and $\bm{b}$ parameters for the x and y coordinates. The final loss is formally defined as:
\begin{equation}
    L_{\mathrm{unc}}=\sum_{i=1}^{2}\mathrm{log}(2b_i)+\frac{|v_i-\mu_i|}{b_i}
    \label{eq:lapalace_loss}
\end{equation}

\subsection{MULTI-RATE GUIDANCE}

In hierarchical frameworks, the high-level planner is typically implemented using a large model, which significantly slows down the overall inference process. For instance, GoalFlow operates at an inference rate of only 12 FPS, significantly lower than the lightweight trajectory decoder of DiffusionDrive (39 FPS), which limits its practical deployment. To address this issue, we propose three lightweight approaches (MLP-Based Predictor, Linear Kinematic Extrapolation, and DAC-Score Guided Kinematic Extrapolation) that allow the high-level module to avoid generating guidance at every frame, thereby enabling multi-rate guidance for the low-level planner. Specifically, at each frame, we predict an extended goal point $\bm{\mu}_\mathrm{extend}$ for the next frame, while directly inheriting the current frame’s uncertainty parameter $\bm{b}$.

\noindent \textbf{\textit{MLP-Based Predictor.}} We first attempt to use a learning-based approach to invoke the high-level module once every two frames. For goal prediction head, we explore two designs: (1) a shared MLP head that jointly predicts goal points for the next two frames, and (2) two separate MLP heads, each predicting one frame with uncertainty guidance. The shared-head design shows unstable training, likely due to differing optimization targets across frames. Therefore, we adopt the separate-head approach. Details are presented in the experiments.

\begin{figure}[h]
  \centering
  \includegraphics[width=0.45\textwidth]{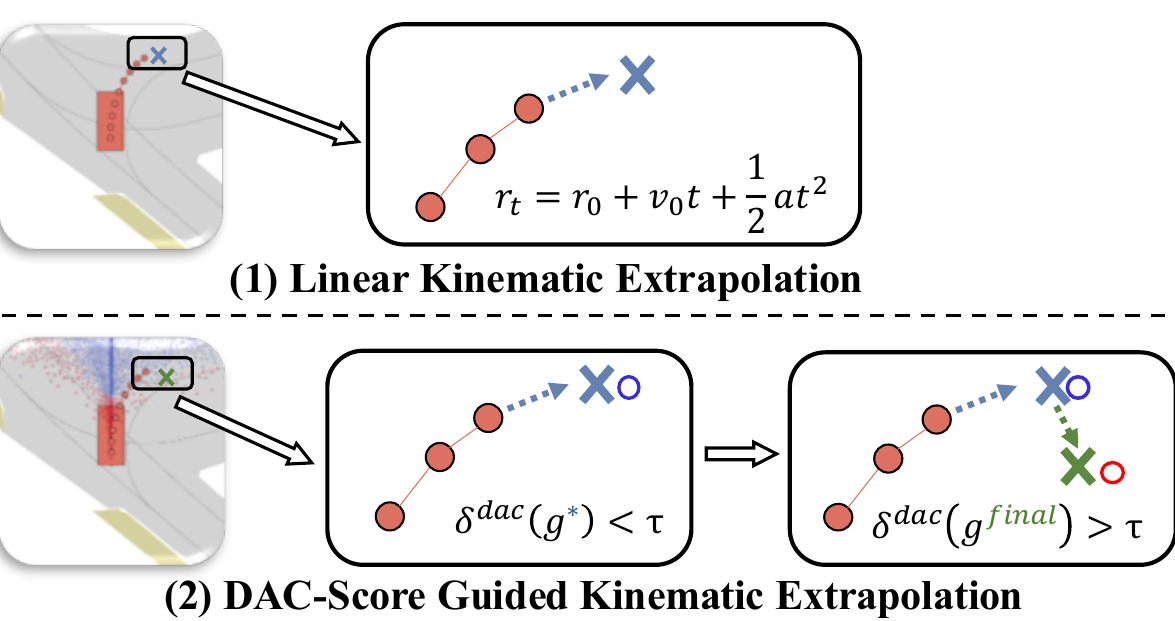}

    \caption{\textbf{Extended Goal Point Extrapolation.} 
    (1) Linear Kinematic Extrapolation directly performs linear extrapolation based on historical trajectory points; 
    (2) DAC-Score Guided Kinematic Extrapolation refines the extended goal point prediction by incorporating DAC scores during extrapolation. 
    In the DAC map visualization, red points indicate higher DAC scores approaching 1, while blue points represent lower scores near 0.}
  \label{fig:extended_goal}
  \vspace{-1.3em}
\end{figure}

\noindent \textbf{\textit{Linear Kinematic Extrapolation.}} We also explore rule-based approaches. A naive interpolation method involves extrapolating the future trajectory using basic laws of motion. Assuming the vehicle undergoes uniform linear motion over a short horizon, we extrapolate the goal point based on the last two points of the current predicted trajectory. This can be formulated using the following physics-based equation:
\begin{equation}
    \bm{\mu}_\mathrm{linear} = \bm{\mu}_\mathrm{current} + \mathbf{v_0}\Delta{t} + \frac{1}{2}\mathbf{a}(\Delta{t})^2 \quad \text{where} \quad \Delta{t} = t - t_0
    \label{eq:linear_motion}
\end{equation}
where $ \mathbf{r} $, $ \mathbf{v} $, and $ \mathbf{a} $ are all two-dimensional vectors in the plane. $ \mathbf{v}_0 $ and $ t_0 $ represent the velocity and timestamp of the last point in the current predicted trajectory.

\noindent \textbf{\textit{DAC-Score Guided Kinematic Extrapolation.}}
As shown in Fig.~\ref{fig:extended_goal}, while simple kinematic extrapolation is effective in most cases due to the linear nature of vehicle motion, it can lead to off-road predictions during turns. To address this, we refine the extrapolated goal point using the DAC score map to encourage drivable area compliance.

In Section~\ref{sec:uncertainty estimation}, each candidate point in the vocabulary $\mathbb{V}$ is associated with a DAC score $\delta^{\mathrm{dac}}$, where a score closer to 1 indicates higher likelihood of being in the drivable area, while a score closer to 0 suggests the point is likely off-road. Given a linearly kinematic extrapolated goal point $\bm{\mu}_{\mathrm{linear}}$, we first locate its nearest neighbor $\bm{\mu}* \in \mathbb{V}$ and approximate its DAC score using $\delta^{\mathrm{dac}}(\bm{\mu}^*)$. If this score exceeds a threshold $\tau$, we set $\bm{\mu}_{\mathrm{extend}} = \bm{\mu}_{\mathrm{linear}}$. Otherwise, we select the most spatially similar candidate from $\mathbb{V}$ whose DAC score exceeds the threshold:

\begin{equation}
\bm{\mu}_{\mathrm{extend}} = 
\begin{cases}
\bm{\mu}_{\mathrm{linear}}, & \text{if } \delta^{\mathrm{dac}}(\bm{\mu}^*) \geq \tau \\[6pt]
\arg\min\limits_{\substack{\bm{\mu}_j \in \mathbb{V} \\ \delta^{\mathrm{dac}}(\bm{\mu}_j) \geq \tau}} 
\| \bm{\mu}_j - \bm{\mu}_{\mathrm{linear}} \|_2, & \text{otherwise}
\end{cases}
\label{eq:extend extra}
\end{equation}

This selection strategy ensures that the extended goal point is not only consistent with physical extrapolation but also satisfies safety constraints by remaining in the drivable area. 

\subsection{GUIDANCE INJECTION MODULE AND DIFFUSION PLANNER}
In this section, the refined goal point $\bm{\mu}$ and its associated uncertainty $\bm{b}$ are injected into a Guidance Injection module, which enhances the trajectory representation. This enriched representation is then passed to the Diffusion Planner to generate the final trajectory.

\begin{figure}[h]
  \centering
  \includegraphics[width=0.44\textwidth]{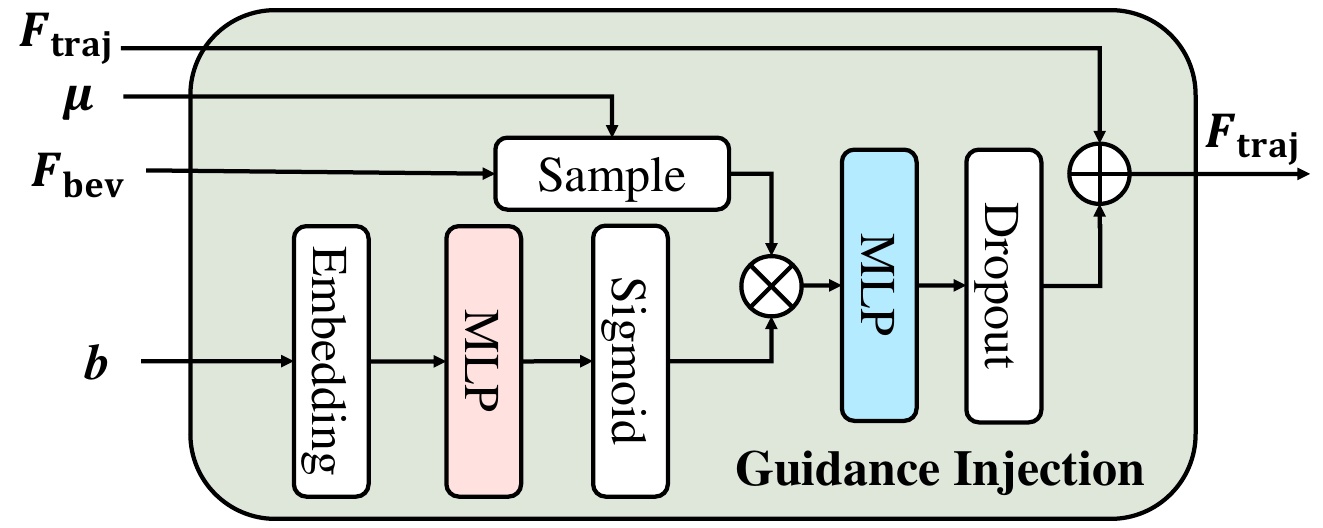}
  \caption{\textbf{Guidance Injection module.}}
  \label{fig:guidance_injection}
\end{figure}

\noindent\textbf{Guidance Injection module.}
As shown in Fig.~\ref{fig:guidance_injection}, the refined goal point $\bm{\mu}$ is first projected from the ego coordinate frame into the BEV space using a projection operator $\Phi$, allowing us to sample the corresponding spatial features from $F_{\mathrm{bev}}$. To reflect the uncertainty in guidance, we compute a confidence weight from $\bm{b}$ using sinusoidal embedding followed by an MLP and a sigmoid activation. The sampled BEV feature is then modulated by this confidence weight through element-wise multiplication to produce feature $F_{\mathrm{guidance}}$:
\begin{equation}
F_{\mathrm{guidance}}=\Phi(F_{\mathrm{bev}},\bm{\mu})\times \mathrm{Sigmoid}(\mathrm{MLP}(\mathrm{Emb}(\bm{b})))
\end{equation}
This guidance is injected into the trajectory query $F_{\mathrm{traj}}$ through residual fusion:
\begin{equation}
F_{\mathrm{traj}}=F_{\mathrm{traj}}+\mathrm{Dropout}(\mathrm{MLP}(F_{\mathrm{guidance}}))
\end{equation}
The trajectory query $F_{\mathrm{traj}}$ is initialized from a set of anchor trajectories, which encode prior knowledge of typical motion patterns and act as carriers for downstream information integration.

\noindent\textbf{Diffusion Planner.}
The final trajectory $\{O^\mathrm{traj}_{1}, ...,O^\mathrm{traj}_{T}\}$ is predicted using a diffusion-based decoder adapted from DiffusionDrive~\cite{liao2025diffusiondrive}. This decoder takes the updated trajectory query $F_{\mathrm{traj}}$ as input and iteratively denoises it to generate the final trajectory.

\vspace{-0.5em}

\section{EXPERIMENT}
\subsection{Dataset and Metrics}
Our method is evaluated on two benchmarks: Navsimv1~\cite{navsimv1} and Navsimv2~\cite{navsimv2}. Navsimv1 uses Navtest as the test set, which includes 134 real-world scenarios totaling over 10,000 frames. All test scenes in Navtest are distributionally similar to the training data, and surrounding agents follow non-reactive replayed logs. In Navsimv1, PDMS is used as the metric, which is a weighted combination of sub-scores, including No at-fault Collisions (NC), where at-fault collisions refer to those that should have been avoided by proper planning, Drivable Area Compliance (DAC), Time-to-collision (TTC), Comfort (Comf.), and Ego Progress (EP).

In contrast, Navsimv2 employs a more challenging test set, Navhard, consisting of 77 scenarios and nearly 6,000 frames, including 450 real and 5,462 synthetic frames. Its evaluation protocol provides a pseudo closed-loop assessment through a two-stage process: In stage 1, it runs a 4-second open-loop simulation using real-world initial scenes without environmental feedback; in stage 2, it evaluates the system in synthesized follow-up scenes initialized from perturbed states, simulating future interactions. The main evaluation metric is EPDMS, which extends the v1 version by incorporating five additional sub-metrics Driving Direction Compliance (DDC), Traffic Light Compliance (TLC), Lane Keeping (LK), History Comfort (HC), and Extended Comfort (EC). These sub-metrics are integrated through a combination of multiplicative factors and additive terms, as detailed in the navsim-v2 documentation\footnote{\url{https://github.com/autonomousvision/navsim/blob/main/docs/metrics.md}}.

\subsection{Implementation Details.} 
For sensor inputs, we concatenate the left-front, front, and right-front camera views into a $256 \times 1024$ image and combine it with LiDAR data in Navtest. As the Navhard includes synthetic scenes without LiDAR, we replace the LiDAR input in these scenarios with a learnable latent feature. The goal point scorer uses the checkpoint of GoalFlow, while the low-level trajectory decoder employs a ResNet-50 backbone. In the DAC-Score Guided Kinematic Extrapolation module, we set the threshold $\tau=0.5$. Defaultly, the slow guidance system performs reasoning once every two frames, while the fast planning system performs reasoning at every frame.

\begin{table}[t] %
    \centering
    \caption{\textbf{\texttt{Navtest} leaderboard of navsimv1.}}
    \resizebox{0.5\textwidth}{!}{
        \begin{tabular}{lcccccc}
            \toprule
            \textbf{Method} & $\mathbf{NC}$ $\uparrow$ & $\mathbf{DAC}$ $\uparrow$ & $\mathbf{TTC}$ $\uparrow$ & $\mathbf{Comf.}$ $\uparrow$ & $\mathbf{EP}$ $\uparrow$ & $\mathbf{PDMS}$ $\uparrow$ \\
            \midrule
            CV & 68.0 & 57.8 & 50.0 & 100 & 19.4 & 20.6 \\
            MLP & 93.0 & 77.3 & 83.6 & 100 & 62.8 & 65.6 \\
            LTF \cite{Chitta2023transfuser} & 97.4 & 92.8 & 92.4 & 100 & 79.0 & 83.8 \\
            TransFuser \cite{Chitta2023transfuser} & 97.7 & 92.8 & 92.8 & 100 & 79.2 & 84.0 \\
            UniAD \cite{hu2023uniad} & 97.8 & 91.9 & 92.9 & 100 & 78.8 & 83.4 \\
            PARA-Drive \cite{weng2024drive} & 97.9 & 92.4 & 93.0 & 99.8 & 79.3 & 84.0 \\
            GoalFlow\ddag~\cite{xing2025goalflow} & {98.2} & {96.4} & {93.8} & {99.4} & {82.6} & {87.9} \\
            Diff.~\cite{liao2025diffusiondrive} & {98.2} & {96.2} & \textbf{94.7} & {100} & {82.2} & {88.1} \\
            Mimir & \textbf{98.2} & \textbf{97.5} & {94.6} & \textbf{100} & \textbf{83.6} & \textbf{89.3} \\
            \bottomrule
        \end{tabular}
    }
    \label{tab:Navtest}
\caption*{\ddag\ denotes version of the trajectory decoder built upon a ResNet backbone with 256-dimensional feature encoding.}
\vspace{-2em}
\end{table}

\begin{table}[t]
\centering
\small
\setlength{\tabcolsep}{1.0pt} 
\caption{\textbf{\texttt{Navhard} leaderboard of navsimv2.}}
\label{tab:sota_transposed}
\begin{tabular}{c c | c c c c c c c} 
    \toprule
    \textbf{Metric} & \textbf{Stage} & \textbf{CV} & \textbf{MLP} & \textbf{LTF~\cite{Chitta2023transfuser}} & {\textbf{GoalFlow\dag~\cite{xing2025goalflow}}} & {\textbf{Diff.\dag~\cite{liao2025diffusiondrive}}} & {\textbf{Mimir}} \\ 
    \midrule
    \multirow{2}{*}{\textbf{NC} $\uparrow$} & S1 & 88.8 & 93.2 & 96.2 & {96.0} & \textbf{96.8} & {95.6}\\
     & S2 & \textbf{83.2} & 77.2 & 77.7 & {79.4} & {80.3} & {80.6}\\
     \cmidrule{2-8}
    \multirow{2}{*}{\textbf{DAC} $\uparrow$} & S1 & 42.8 & 55.7 & 79.5 & \textbf{92.6} & {88.2} & {92.2}\\
     & S2 & 59.1 & 51.9 & {70.2} & \textbf{78.2} & {74.4} & {77.1} \\
     \cmidrule{2-8}
    \multirow{2}{*}{\textbf{DDC} $\uparrow$} & S1 & 70.6 & 86.6 & 99.1 & {99.3} & {99.3} & \textbf{99.7}\\ 
     & S2 & 76.5 & 74.4 & {84.2} & {86.4} & {86.1} & \textbf{89.3} \\
     \cmidrule{2-8}
    \multirow{2}{*}{\textbf{TLC} $\uparrow$} & S1 & 99.3 & 99.3 & 99.5 & {99.3} & {99.3} & \textbf{99.5}\\
     & S2 & 98.0 & {98.2} & 98.0 & {97.7} & \textbf{98.4} & {97.7} \\
    \midrule
    \multirow{2}{*}{\textbf{EP} $\uparrow$} & S1 & 77.5 & 81.2 & 84.1 & {84.0} & \textbf{84.5} & {84.0}\\
     & S2 & 71.3 & 77.1 & {85.1} & {86.5} & \textbf{87.4} & {86.4} \\
     \cmidrule{2-8}
    \multirow{2}{*}{\textbf{TTC} $\uparrow$} & S1 & 87.3 & 92.2 & {95.1} & \textbf{95.7} & {94.6} & {94.6}\\
     & S2 & {81.1} & 75.0 & 75.6 & {76.0} & {76.9} & \textbf{77.3} \\
     \cmidrule{2-8}
    \multirow{2}{*}{\textbf{LK} $\uparrow$} & S1 & 78.6 & 83.5 & {94.2} & {97.1} & {95.5} & \textbf{98.0}\\
     & S2 & {47.9} & 40.8 & 45.4 & {45.5} & \textbf{50.4} & {48.8} \\
     \cmidrule{2-8}
    \multirow{2}{*}{\textbf{HC} $\uparrow$} & S1 & 97.1 & {97.5} & {97.5} & {97.5} & {97.5} & \textbf{97.5}\\
     & S2 & 97.1 & \textbf{97.8} & 95.7 & {94.4} & {95.5} & {94.5}\\
     \cmidrule{2-8}
    \multirow{2}{*}{\textbf{EC} $\uparrow$} & S1 & 60.4 & 77.7 & {79.1} & {40.4} & \textbf{79.1} & {78.2}\\
     & S2 & 61.9 & {79.8} & 75.9 & {40.4} & \textbf{69.4} & {64.0} \\
    \midrule
    \multicolumn{2}{c|}{\textbf{EPDMS} $\uparrow$}  & 10.9 & 12.7 & {23.1} & {28.7} & {28.9} & \textbf{34.6}\\ 

    \bottomrule
    \label{tab:Navhard}

\end{tabular}
\caption*{\dag\ indicates the version trained using open-source code without LiDAR input.}
\vspace{-2.5em}
\end{table}

\begin{table}[h]
\centering
\setlength{\tabcolsep}{3.0pt} 
\caption{\textbf{FPS of SOTA methods on \texttt{Navhard}}.}
\begin{tabular}{c|c|c c c}
    \toprule
     & \textbf{Module} & \textbf{DiffusionDrive\dag~\cite{liao2025diffusiondrive}} & \textbf{GoalFlow\dag~\cite{xing2025goalflow}} & \textbf{Mimir} \\
    \midrule
    \multirow{2}{*}{\textbf{FPS}$\uparrow$} 
    & high-level & / & 13 & \textbf{21} \\
    & low-level  & \textbf{39} & 15 & 36 \\
    \bottomrule
\end{tabular}
\label{tab:FPS}
\end{table}

For training, the goal point module is supervised using the uncertainty-aware loss $L_{\mathrm{unc}}$ introduced in Section~\ref{sec:uncertainty estimation}, and the trajectory decoder is trained with an L1 loss. All models are trained on 8 NVIDIA H20 GPUs using the navtrain with a batch size of 64. We use a hidden dimension of 256 and a learning rate of $6 \times 10^{-4}$. No external datasets, ensembled models, or test-time augmentations are used.

\begin{table}[t]
\centering
\setlength{\tabcolsep}{8pt} 
\caption{\textbf{Ablation of Different Components.}}
\label{tab:sota_transposed}
\begin{tabular}{c c | c c c c} 
    \toprule
    \textbf{Metric} & \textbf{Stage} & $\mathcal{M}_0$ & $\mathcal{M}_1$ & $\mathcal{M}_2$ & $\mathcal{M}_3$ \\ 
    \midrule
    \multirow{2}{*}{\textbf{NC} $\uparrow$} & S1 & \textbf{96.8} & 95.6 & {96.4} & {96.2}\\
     & S2 & \textbf{80.3} & 78.9 & {79.6} & {79.6}\\
     \cmidrule{2-6}
    \multirow{2}{*}{\textbf{DAC} $\uparrow$} & S1 & 88.2 & \textbf{92.0} & {90.4} & {90.6}\\
     & S2 & 74.4 & \textbf{78.6} & {77.6} & {78.5}\\
     \cmidrule{2-6}
    \multirow{2}{*}{\textbf{DDC} $\uparrow$} & S1 & 99.3 & 100 & {99.3} & \textbf{100}\\ 
     & S2 & 86.1 & 88.0 & {88.2} & \textbf{88.8}\\
     \cmidrule{2-6}
    \multirow{2}{*}{\textbf{TLC} $\uparrow$} & S1 & 99.3 & 99.3 & {99.3} & \textbf{99.3}\\
     & S2 & \textbf{98.4} & {97.6} & {97.8} & {97.5}\\
    \midrule
    \multirow{2}{*}{\textbf{EP} $\uparrow$} & S1 & 84.5 & \textbf{84.9} & {84.4} & {84.2}\\
     & S2 & \textbf{87.4} & 86.4 & {86.3} & {86.0}\\
     \cmidrule{2-6}
    \multirow{2}{*}{\textbf{TTC} $\uparrow$} & S1 & 94.6 & 95.1 & {95.1} & \textbf{95.3}\\
     & S2 & {76.9} & 75.9 & {76.7} & \textbf{77.0}\\
     \cmidrule{2-6}
    \multirow{2}{*}{\textbf{LK} $\uparrow$} & S1 & 95.5 & \textbf{98.0} & {97.1} & {97.5}\\
     & S2 & \textbf{50.4} & 47.2 & {48.2} & {48.5}\\
     \cmidrule{2-6}
    \multirow{2}{*}{\textbf{HC} $\uparrow$} & S1 & 97.5 & {97.7} & {97.7} & \textbf{97.7}\\
     & S2 & \textbf{95.5} & {94.6} & {93.8} & {94.2}\\
     \cmidrule{2-6}
    \multirow{2}{*}{\textbf{EC} $\uparrow$} & S1 & 79.1 & 70.2 & {77.7} & \textbf{78.2}\\
     & S2 & \textbf{69.4} & {53.7} & {59.8} & {64.1}\\
    \midrule
    \multicolumn{2}{c|}{\textbf{EPDMS} $\uparrow$}  & 28.9 & 31.1 & {31.1} & \textbf{33.3}\\ 
    \bottomrule
    \label{tab:ablation of component}
\end{tabular}
\caption*{
$\mathcal{M}_0$ denotes the DiffusionDrive model;
$\mathcal{M}_1$ denotes the addition of goal point injection;
$\mathcal{M}_2$ denotes $\mathcal{M}_1$ with Gaussian uncertainty modeling;
$\mathcal{M}_3$ denotes $\mathcal{M}_1$ with Laplace uncertainty modeling.
}
\vspace{-2.5em}
\end{table}
\vspace{-6pt}

\subsection{Main Result.}
In Table~\ref{tab:Navtest}, we compare our method, Mimir, with several state-of-the-art approaches on the Navtest benchmark, which primarily consists of real-world driving scenarios. To ensure a fair comparison, we use trajectory decoders of the same capacity across all methods to ensure a fair comparison.: a Resnet-50 backbone with 256-dimensional feature representations. Mimir presents a strong performance, achieving the best scores in DAC and EP, which suggests that our method generalizes well to regular in-distribution scenes. Furthermore, it demonstrates the ability to improve driving efficiency (high EP score) without compromising safety.

As shown in Table~\ref{tab:Navhard}, we report results on the more challenging benchmark Navhard, including a large portion of synthetically generated scenarios.  Our method achieves nearly a 20\% improvement over the best baseline in terms of overall EPDMS and ranks first in several sub-metrics. Compared to DiffusionDrive, Mimir achieves consistently higher scores across key safety and compliance metrics such as DAC, and DDC with gains of 3–4 points. We attribute this to the hierarchical structure of our model, where a strong high-level module provides more accurate guidance for downstream trajectory generation. Interestingly, for critical metrics like DDC, Mimir performs better on stage 2 synthetic scenes than on Stage 1 real scenes. This suggests that our model is robust to distribution shifts and better equipped to handle difficult situations. Furthermore, compared to the similar hierarchical framework GoalFlow, Mimir achieves a significantly higher score in EC, which measures the consistency between adjacent frames in the generated trajectory, indicating smoother and more stable driving behavior. 

In Table~\ref{tab:FPS}, we compare the FPS of different modules across methods. Compared to GoalFlow, our high-level module achieves a 1.6$\times$ speedup, owing to the design of our multi-rate guidance mechanism. Moreover, by adopting a truncated diffusion process inherited from DiffusionDrive, our low-level module achieves faster inference with fewer denoising steps compared to GoalFlow (2 vs. 5), achieving higher FPS (32 vs. 15).

\begin{table}[t]
\centering
\setlength{\tabcolsep}{7pt} 
\caption{\textbf{Extended Goal Point Prediction.}}
\label{tab:sota_transposed}
\begin{tabular}{c c | c c c c c} 
    \toprule
    \textbf{Metric} & \textbf{Stage} & $\Gamma_0$ & $\Gamma_1$ & $\Gamma_2$ & $\Gamma_3$ & $\Gamma_4$ \\ 
    \midrule
    \textbf{Latency (ms)}$\downarrow$ & / & \textcolor{gray}{3.09} & 3.61 & \textbf{3.27} & 3.62 & 4.13\\
    \midrule
    \multirow{2}{*}{\textbf{NC} $\uparrow$} & S1 & \textbf{96.2} & 95.8 & 95.6 & 95.6 & 95.6\\
     & S2 & {79.6} & \textbf{81.4} & {80.5} & {80.6} & 80.7\\
     \cmidrule{2-7}
    \multirow{2}{*}{\textbf{DAC} $\uparrow$} & S1 & 90.6 & 91.5 & 90.8 & \textbf{92.2} & 92.2 \\
     & S2 & \textbf{78.5} & 78.0 & {77.1} & {77.1} & 76.9\\
     \cmidrule{2-7}
    \multirow{2}{*}{\textbf{DDC} $\uparrow$} & S1 & \textbf{100} & 99.6 & 99.5 & 99.7 & blue{99.6}\\ 
     & S2 & 88.8 & 88.8 & {88.0} & \textbf{89.3} & 89.2 \\
     \cmidrule{2-7}
    \multirow{2}{*}{\textbf{TLC} $\uparrow$} & S1 & 99.3 & 99.3 & 99.5 & \textbf{99.5} & 99.5\\
     & S2 & 97.5 & \textbf{97.8} & {97.6} & {97.7} & 97.7\\
    \midrule
    \multirow{2}{*}{\textbf{EP} $\uparrow$} & S1 & 84.2 & 84.1 & \textbf{84.5} & 84.0 & 84.0\\
     & S2 & 86.0 & 86.3 & \textbf{87.1} & {86.4} & 86.2\\
     \cmidrule{2-7}
    \multirow{2}{*}{\textbf{TTC} $\uparrow$} & S1 & \textbf{95.3} & {95.3} & 94.2 & {94.6} & 94.4\\
     & S2 & {77.0} & 77.6 & {77.0} & \textbf{77.3} & 77.3\\
     \cmidrule{2-7}
    \multirow{2}{*}{\textbf{LK} $\uparrow$} & S1 & 97.5 & 98.0 & 98.0 & \textbf{98.0} & 97.9\\
     & S2 & {48.5} & \textbf{49.0} & {48.3} & {48.8} & 49.0 \\
     \cmidrule{2-7}
    \multirow{2}{*}{\textbf{HC} $\uparrow$} & S1 & \textbf{97.7} & {97.7} & {97.7} & {97.5} & 97.5\\
     & S2 & 94.2 & {94.4} & {94.4} & \textbf{94.5} & \textbf{94.6}\\
     \cmidrule{2-7}
    \multirow{2}{*}{\textbf{EC} $\uparrow$} & S1 & 78.2 & 78.2 & \textbf{79.1} & {78.2} & 78.2\\
     & S2 & 64.1 & {63.0} & \textbf{64.3} & {64.0} & 63.2\\
    \midrule
    \multicolumn{2}{c|}{\textbf{EPDMS} $\uparrow$} & 33.3 & 33.5 & {34.0} & \textbf{34.6} & 34.4\\ 
    \bottomrule
    \label{tab:ablation of interpolation}
\end{tabular}
\caption*{$\Gamma_0$ serves as our baseline (equivalent to $\mathcal{M}_3$ without extension); 
$\Gamma_1$ employs an MLP-based predictor; \
$\Gamma_2$ implements linear kinematic extrapolation;
$\Gamma_3$ enhances the kinematic extrapolation with DAC score; and
$\Gamma_4$ further extends $\Gamma_3$ by predicting the next-step uncertainty via an additional MLP head.}

\vspace{-2.5em}
\end{table}
\vspace{-6pt}

\begin{figure*}[ht]
  \centering
  \includegraphics[width=1.0\textwidth]{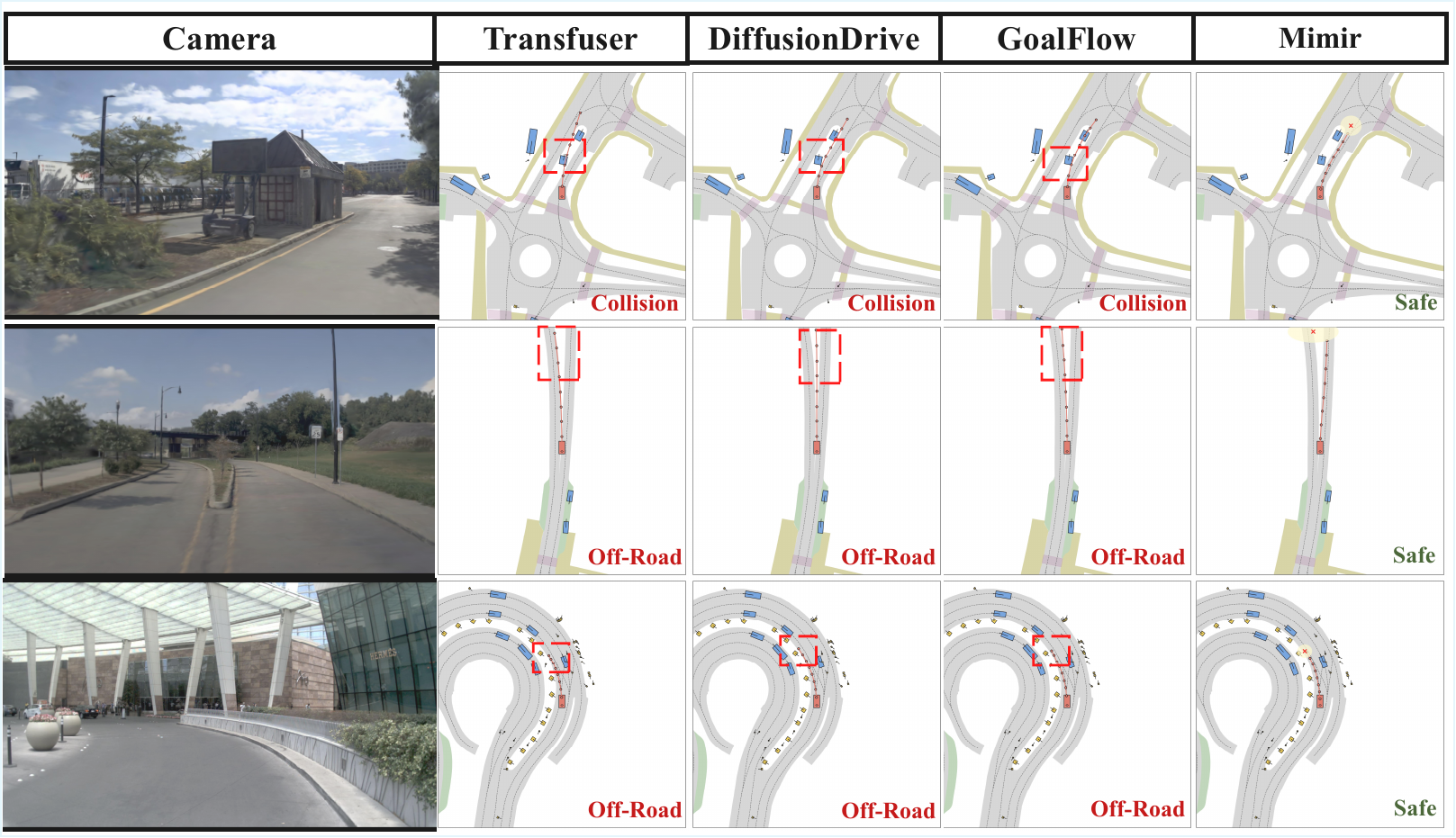}
  \caption{\textbf{Visualization.} From top to bottom, we illustrate the capabilities of the four algorithms in Exit-Ramp Scenario, forked-intersection scenario, and roundabout scenario, respectively. The red cross represents the predicted goal point and the size of the yellow area around the goal point represents the level of uncertainty. Mimir leverages uncertainty estimation to mitigate the effects of inaccurate high-level guidance, enabling the generation of safer trajectories.}
  \label{fig:visual}
\vspace{-2.0em}
\end{figure*}

\subsection{Ablation of Different Components.}
In this section, we perform an ablation study to examine the impact of individual components in our model, as shown in Table~\ref{tab:ablation of component}. Compared to $\mathcal{M}_0$, $\mathcal{M}_1$ yields notable improvements in key metrics such as DAC and DDC, suggesting that the additional information effectively enhances spatial guidance in the low-level trajectory decoder. However, we also observe a considerable drop in the EC score compared to $\mathcal{M}_0$, which we attribute to a common limitation in hierarchical frameworks like GoalFlow, where the top and bottom modules are not optimized in a fully end-to-end manner.

Compared to $\mathcal{M}_1$, $\mathcal{M}_2$ achieves a similar EPDMS score but suffers from training instability. We believe this is due to the quadratic penalty in the log-likelihood of Gaussian distributions, which makes them more sensitive to outliers than the linear penalty in Laplace distributions.

In contrast, $\mathcal{M}_3$ shows clear improvements over $\mathcal{M}_1$, with a 2-point gain in overall EPDMS and significant improvements in TTC and EC. Notably, TTC estimates the likelihood of collision within a short time horizon. We hypothesize that uncertainty modeling in $\mathcal{M}_3$ helps the model capture potential interaction risks with nearby agents, thereby reducing collision probability. Additionally, the improved EC score compared to both $\mathcal{M}_1$ and $\mathcal{M}_2$ suggests that modeling uncertainty in goal point predictions provides temporal tolerance for slight fluctuations between adjacent frames.

\subsection{Extended Goal Point Prediction.}
A comparative study, as shown in Table~\ref{tab:ablation of interpolation}, focuses on single extended goal point prediction, due to the constraint of at most two adjacent frames in the Navhard benchmark. We analyze four approaches using a single NVIDIA H20 GPU. Our latency analysis, which measures the combined inference time of the uncertainty estimation and multi-rate guidance modules, reveals that $\Gamma_1$ introduces minimal computational overhead compared to other methods, while $\Gamma_2$ and $\Gamma_3$ exhibit similar latency levels. Notably, all variants maintain sub-10ms execution times, demonstrating their practical feasibility for real-time systems.

Despite $\Gamma_1$'s lightweight design, it achieves comparable performance to $\Gamma_0$ in terms of overall EPDMS. We believe this is because the temporal gap between two consecutive frames is very small, resulting in highly similar condition features between them.

For $\Gamma_2$, this simple rule-based method yields strong performance surprisingly. We attribute this to the fact that, over short time horizons (e.g., 0.5 seconds), vehicle motion tends to follow linear kinematics. 

$\Gamma_3$ achieves a higher EPDMS, with a particularly notable improvement of 1.4 points in the Stage 1 DAC score. This highlights the effectiveness of DAC-guided replanning. However, we observe that both $\Gamma_2$ and $\Gamma_3$ perform similarly on Stage 2 DAC. We hypothesize that this is due to prediction errors in the DAC score map within the synthetic data.

$\Gamma_4$ shows that the performance remains comparable to the $\Gamma_3$. We speculate that this is because the prediction horizon is relatively short (within 0.5s), where the fluctuation range of uncertainty is small, leading to limited differences between the two strategies.

\vspace{-0.5em}
\begin{table}[h]
\centering
\textbf{Table VI. PDMS on \texttt{Navtest} under different extrapolation horizons.}\\[4pt]
\begin{tabular}{llccccc}
\toprule
\textbf{n} & \textbf{Method} & \textbf{0} & \textbf{3} & \textbf{5} & \textbf{10} & \textbf{20} \\
\midrule
\multirow{2}{*}{\textbf{PDMS $\uparrow$}}
& \textbf{L-KE}   & 89.3 & 82.2 & 55.0 & 36.2 & 26.6 \\
& \textbf{DAC-KE} & {89.3} & \textbf{86.1} & \textbf{65.1} & \textbf{45.7} & \textbf{43.2} \\
\bottomrule
\end{tabular}
\label{tab:Navtest_pdms}
\vspace{-6pt}
\end{table}
In Tab. VI, we further assess both Linear Kinematic Extrapolation (L-KE) and DAC-Score Guided Kinematic Extrapolation (DAC-KE) on Navtest, which provides continuous history frames suitable for long-horizon analysis. In this setting, the goal points are extrapolated from the $n$-th historical frame, while $n=0$ denotes using the goal point from the current frame. We find DAC-KE effectively alleviates the limitations of the L-KE and its practical extrapolation limit is around 3.

\subsection{Quality Results.}
In Fig.~\ref{fig:visual}, we present a visual comparison between Mimir and four other methods. Compared to the baselines, Mimir consistently generates safer trajectories. We observe that when the goal point is inaccurate, Mimir exhibits higher uncertainty, which serves as a safeguard against low-quality high-level guidance. On the other hand, when the goal point is relatively accurate, our trajectory planner produces higher-quality trajectories. We attribute this to our guidance injection module, which effectively integrates environmental context with high-level guidance for improved planning.

\section{CONCLUSIONS AND FUTURE WORK}
We present Mimir, a hierarchical dual-system framework for end-to-end autonomous driving with integrated uncertainty-aware guidance. Goal point uncertainty is modeled via a Laplace distribution to improve robustness. A DAC-guided kinematic extrapolation enables fast extended goal prediction for multi-rate processing, while a Guidance Injection module effectively integrates high-level guidance into the trajectory planner. Experiments on Navhard and Navtest show SOTA performance, with ablation studies confirming the effectiveness of each component.

Looking forward, we plan to extend our work to VLA by exploring how uncertainty in language models can influence the performance of low-level planners. We will also continue to push the performance ceiling of the guidance extrapolation method on longer time horizons, as well as develop a goal-caching mechanism to enable asynchronous communication between hierarchical modules.

\ifCLASSOPTIONcaptionsoff
  \newpage
\fi

\end{document}